\documentclass[conference]{IEEEtran}
\IEEEoverridecommandlockouts
\usepackage{cite}
\usepackage{amsmath,amssymb,amsfonts}
\usepackage{graphicx, color}
\usepackage{subcaption}
\usepackage[export]{adjustbox}
\usepackage{textcomp}
\usepackage{url} 
\usepackage[table]{xcolor} 
\usepackage{booktabs} 
\usepackage{paralist}
\usepackage{enumitem}
\usepackage{multirow}
\usepackage{hyperref}
\usepackage{algorithmic}
\usepackage[linesnumbered,ruled,vlined]{algorithm2e}
\hypersetup{
    colorlinks=true,
    linkcolor=red, 
    citecolor=blue, 
    urlcolor=magenta    
}
\def\ie{{\em i.e.}}
\def\eg{{\em e.g.}}
\def\BibTeX{{\rm B\kern-.05em{\sc i\kern-.025em b}\kern-.08em
    T\kern-.1667em\lower.7ex\hbox{E}\kern-.125emX}}

\newcommand{\smallurl}[1]{\footnotesize\url{#1}}

\definecolor{baselinecolor}{gray}{.9}

\begin{document}

\title{Robust AI-Generated Face Detection with Imbalanced Data}

\author{\IEEEauthorblockN{Yamini Sri Krubha\textsuperscript{1}$^+$\thanks{$^+$Co-first Authors}, Aryana Hou\textsuperscript{2}$^+$,  Braden Vester\textsuperscript{1}$^+$, Web Walker\textsuperscript{1}$^+$, Xin Wang\textsuperscript{3}, Li Lin\textsuperscript{1}, Shu Hu\textsuperscript{1}$^*$\thanks{$^*$Corresponding Author} }
\IEEEauthorblockA{
{\textsuperscript{1}Purdue University, {\tt \small \{ykrubha, bjvester, wiwalker, lin1785,  hu968\}@purdue.edu} }\\
\textsuperscript{2}Clarkstown High School South, {\tt \small aryanahou@gmail.com}\\
\textsuperscript{3}University at Albany, State University of New York, {\tt \small xwang56@albany.edu}}
}

\maketitle
\thispagestyle{plain}
\pagestyle{plain}

\begin{abstract}
Deepfakes, created using advanced AI techniques such as Variational Autoencoder and Generative Adversarial Networks, have evolved from research and entertainment applications into tools for malicious activities, posing significant threats to digital trust. 
Current deepfake detection techniques have evolved from CNN-based methods focused on local artifacts to more advanced approaches using vision transformers and multimodal models like CLIP, which capture global anomalies and improve cross-domain generalization. Despite recent progress, state-of-the-art deepfake detectors still face major challenges in handling distribution shifts from emerging generative models and addressing severe class imbalance between authentic and fake samples in deepfake datasets , which limits their robustness and detection accuracy. To address these challenges, we propose a framework that combines dynamic loss reweighting and ranking-based optimization, which achieves superior generalization and performance under imbalanced dataset conditions. The code is available at \url{https://github.com/Purdue-M2/SP_CUP}.

\end{abstract}


\section{Introduction}

Deepfakes represent one of the most concerning technological advancements of the digital age, leveraging artificial intelligence (AI) to create hyper-realistic yet entirely fabricated media \cite{lin2025ai,chen2024self,zheng2024few,lin2024detecting,chen2023harnessing,guo2022open}. These synthetic videos and images are typically generated using deep learning techniques such as Variational Autoencoder (VAE) \cite{fu2024vb} and  Generative Adversarial Networks (GANs) \cite{goodfellow2014generative,zhang2023x,wang2023gan}, enabling seamless face-swapping or voice replication. Originally developed for entertainment and academic research, deepfakes have since become a tool for malicious activities, including disinformation campaigns, identity fraud, and privacy violations \cite{chesney2019deep}. Researchers warn that the democratization of deepfake tools threatens to erode public trust in digital media \cite{vaccari2020deepfakes}, while advances in detection methods struggle to keep pace with increasingly sophisticated manipulations \cite{nguyen2022deep}.


In real-world scenarios, deepfake datasets are heavily imbalanced, often containing far fewer fake samples compared to authentic ones \cite{li2018exposing}. This imbalance arises because genuine images and videos are abundant, while high-quality deepfake samples are comparatively scarce and costly to produce. As a result, deepfake detectors trained on such skewed datasets tend to become biased toward the majority (real) class, leading to poor generalization and significantly degraded performance when identifying minority-class (fake) samples \cite{santosh2024robust, pu2022learning}. Moreover, traditional data balancing techniques such as oversampling or undersampling are often insufficient, as they can introduce redundancy or discard important diversity in the data \cite{chawla2002smote}. These challenges highlight the need for specialized strategies that can address the class imbalance in a way that preserves feature diversity and enhances the model's robustness to both common and unseen manipulations.

Current deepfake detection techniques largely depend on deep learning models. Initial methods predominantly used convolutional neural networks (CNNs) to identify localized artifacts and inconsistencies typical of synthetic content \cite{afchar2018mesonet}. In contrast, recent advancements have embraced vision transformers (ViTs) \cite{dosovitskiy2020image}, which are particularly effective at capturing long-range interactions and subtle, global anomalies often found in AI-generated visuals. A notable progression in this field is the adoption of multimodal approaches, such as those based on CLIP \cite{radford2021learning}, which leverage aligned image-text representations to enhance cross-domain robustness. For instance, \cite{cozzolino2024raising} demonstrated that CLIP’s pretrained representations can effectively distinguish deepfakes across diverse generative models by capturing semantically meaningful features. 


Despite these advances, state-of-the-art detectors face two critical challenges. First, their robustness to distribution shifts remains inadequate, particularly when analyzing outputs from emerging architectures like latent diffusion models \cite{rombach2022high,huang2025diffusion,chen2024masked}. Second, they struggle with data imbalance \cite{huang2024robustly}, as real-world datasets often contain significantly fewer fake samples than genuine ones, leading to biased classifiers that prioritize majority-class accuracy at the expense of minority-class detection \cite{santosh2024robust}. While some studies address imbalance via resampling \cite{chawla2002smote} or focal loss variants \cite{lin2017focal}, these methods fail to account for the high-dimensional feature-space variability of deepfakes. 

To address the challenges of generalization and robustness in deepfake detection, we propose a novel framework that integrates dynamic loss reweighting and enhanced ranking optimization. Our approach dynamically adjusts the loss contributions of minority (fake) and majority (real) samples during training, ensuring balanced learning across classes. Additionally, we introduce a ranking-based optimization mechanism that prioritizes discriminative features for improved generalization across diverse deepfake generation techniques, including VAEs and GANs. Through extensive experiments on benchmark datasets, we demonstrate that our method significantly outperforms existing approaches in both balanced and imbalanced data scenarios while maintaining robustness against adversarial perturbations and distribution shifts. The proposed solution provides a principled way to handle real-world deepfake detection where data imbalance and evolving generative techniques are critical concerns.
Our contributions are summarized as follows:
\begin{enumerate}
    \item We proposed a novel method to improve the robustness and generalization of AI-generated face detection based on the foundation model CLIP.
    \item Our method outperforms the baselines, as demonstrated in experiments on the datasets provided by the EEE SPS Signal Processing Cup 2025.
\end{enumerate}

\section{Related Work}
\subsection{Deepfake Generation} 

Deepfake generation has evolved significantly over the years, incorporating multiple machine-learning techniques to synthesize highly realistic fake media.
Traditional methods primarily relied on image manipulation techniques such as face swapping and retouching. While effective to some extent, these techniques were labor-intensive and lacked the ability to produce seamless results, often leaving visible artifacts that could be easily detected\cite{goodfellow2014generative,fan2024synthesizing}. With the advent of Generative Adversarial Networks (GANs) \cite{goodfellow2014generative}, the quality and realism of deepfakes improved dramatically. GANs have been widely adopted for generating realistic facial images and videos, with notable advancements such as StyleGAN \cite{karras2019style}, which introduced style-based generators to improve the quality and diversity of generated images. More recently, diffusion models have emerged as a powerful alternative, offering high-quality image synthesis with better control over the generation process \cite{ho2020denoising}. These models have demonstrated superior performance in generating high-quality images with fine-grained details and better control over the generation process.

\subsection{Deepfake Detection}

Deepfake detection methods can be broadly categorized into naive, spatial-based, and frequency-based approaches. Naive methods \cite{guo2022eyes,hu2021exposing} often rely on basic image features or simple classifiers but are limited in their ability to handle sophisticated generative models. Spatial-based methods \cite{yang2023crossdf,fan2023attacking} focus on extracting intrinsic image features such as texture, gradients, and local dimensionality. \cite{zhong2023rich} utilize texture contrast between high- and low-detail regions, emphasizing the challenges generative models face in replicating rich textures. \cite{nguyen2023unmasking} leverage gradient-based features to differentiate synthetic images from real ones, while \cite{lorenz2023detecting} introduce multi Local Intrinsic Dimensionality (multiLID), which employs ResNet18 to extract low-dimensional texture representations for detection, though its effectiveness across different datasets remains limited\cite{lin2024detecting}.

On the other hand, frequency-based methods analyze image frequency components to expose artifacts invisible in the spatial domain. \cite{wolter2022wavelet} propose a multi-scale wavelet representation that merges spatial and frequency features but find minimal advantages in higher-order wavelets.\cite{xi2023ai} develop a dual-stream network that integrates texture and low-frequency analysis for improved detection across varying resolutions. \cite{poredi2023ausome} utilizes spectral comparisons of Discrete Fourier and Cosine Transforms to identify DALL-E 2 images, while \cite{bammey2023synthbuster} applies a cross-difference filter to highlight frequency-based artifacts, enhancing generalization and robustness against compression. These methods collectively advance deepfake detection by exploiting spatial inconsistencies and frequency anomalies in AI-generated images\cite{lin2024detecting}.

\subsection{Robust Deepfake Detection}


Deepfake detection remains a challenging task, particularly due to the class imbalance in real-world datasets \cite{guo2022robust}, where manipulated images or videos are significantly underrepresented compared to genuine samples. \cite{pu2022learning} proposes a Deep Dual-Level Network (DDLN) that addresses this issue by leveraging both global and local feature representations to enhance detection robustness. Their model integrates an attention mechanism to emphasize discriminative features in manipulated regions, effectively mitigating the impact of imbalanced training data and demonstrating improved generalization across multiple datasets. Building on the challenge of imbalance, \cite{lin2024preserving} highlights the issue of fairness generalization, showing that deepfake detection models often exhibit inconsistent performance across diverse demographic groups due to biased training data. They introduce techniques to preserve fairness during training, ensuring more equitable outcomes. Similarly, \cite{ju2024improving} focuses on improving fairness through adversarial training and data augmentation, which balance underrepresented classes and enhance the generalizability of detection models. Collectively, these studies emphasize the importance of addressing data imbalance and fairness to build robust and equitable deepfake detection systems.

\section{Method} 
\subsection{Overview}
\begin{figure*}[t]
    \centering
    \includegraphics[width=1\textwidth]{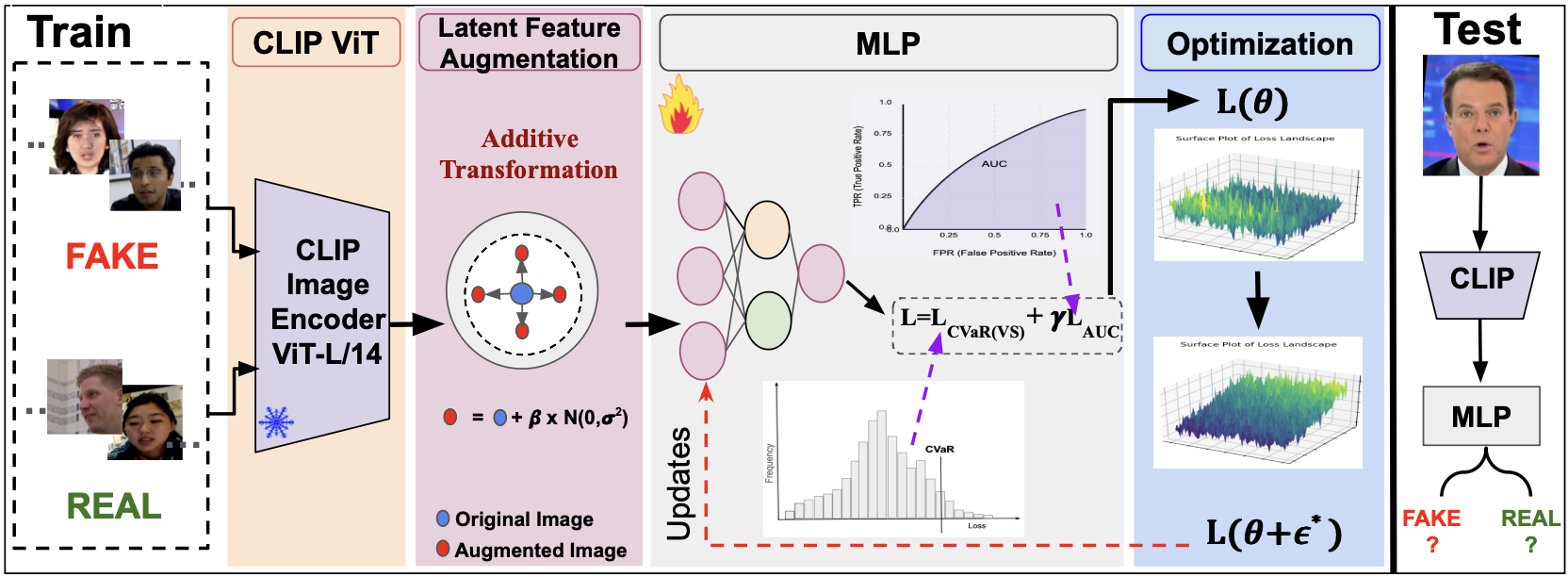}
    \vspace{-5mm}
    \caption{\textit{Overview of our proposed model using CLIP ViT-L/14 for encoding input images and an MLP module trained with a robust loss function combining CVaR on visual similarity and weighted AUC. The optimization process leverages a flattened loss landscape to ensure robust deepfake detection between real and fake images.}}
    \vspace{-4mm}
    \label{fig:SL}
\end{figure*}

Figure \ref{fig:SL} provides an overview of our proposed framework.
The model architecture consists of CLIP ViT L/14 (frozen) combined with three fully connected layers (MLP) that are trainable. To enhance generalization, latent feature augmentation is applied by introducing controlled noise to extracted features using an additive transformation. The MLP with Batch Normalization, ReLU activation, and Dropout, refines these features, while optimization is guided by a composite loss function integrating Conditional Value at Risk (CVaR) on visual similarity and a weighted AUC loss. CVaR focuses on hard-to-classify samples by prioritizing worst-case performance, while AUC loss improves classification accuracy. Additionally, the loss landscape is flattened to enhance robustness against adversarial perturbations and overfitting. During inference, the trained model processes test images through the CLIP encoder and MLP, determining whether an image is real or fake with high accuracy. This architecture effectively enhances deepfake detection by leveraging pre-trained visual features, robust augmentation techniques, and tailored loss functions.
\subsection{Pre-trained Models}
The pre-trained model used in our deepfake detection system is the OpenAI CLIP ViT-L/14 model\cite{Ilharco_OpenCLIP_2021}. CLIP (Contrastive Language-Image Pre-training)\cite{radford2021learning} is a pre-training model developed by OpenAI to convert text supervision to vision models. It is trained on an extensive dataset of 400 million image-text pairs. This large-scale pre-training allows CLIP to capture high-level visual representations that are generalizable across various domains, including facial image analysis. The ViT-L/14 variant employs a Vision Transformer architecture with a smaller starting patch size of 14x14, which is particularly beneficial for capturing fine-grained details in facial images that are crucial for deepfake detection\cite{lin2024robust}.

The integration of CLIP ViT-L/14 into our deepfake detection workflow is designed for effectiveness and efficiency. The image encoder of CLIP ViT-L/14 \cite{Ilharco_OpenCLIP_2021} is used as a frozen feature extractor, mapping input facial images to 768-dimensional feature representations. These high-level features are then processed for deepfake detection. This approach leverages transfer learning, allowing our model to benefit from CLIP's rich understanding of visual features even with limited deepfake-specific training data. By using a pre-trained CLIP model, we can effectively extract relevant features for deepfake detection while minimizing the need for extensive training on large-scale datasets.

\subsection{Training Data Augmentation}
Given a dataset $\mathcal{D}= \{(X_i, Y_i)\}_{i=1}^n$ with size $n$, where $X_i$ is the $i$-th image and $Y_i \in \{0,1\}$ is the $i$-th sample label (0 means fake, 1 means real). Feed CLIP visual encoder with dataset $\mathcal{D}$, and use its final layer to map the training data to their feature representations (of 768 dimensions). We get the resulting feature bank $\mathcal{C} = \{(F_i, Y_i)\}_{i=1}^n$ and further use the feature bank, which is our training set, to design a 3-layer MLP classifier with a classification head $h$. In order to let the MLP classifier learn a comprehensive representation of the forgery feature and a more robust decision boundary, we enlarge the feature space by interpolating samples before feeding the features into the MLP classifier. Specifically, we apply a traditional and effective augmentation, additive transformation. By adding random noise, for example, Gaussian Mixture Model moise with zero mean, $F_i$ can be perturbed with the scaling factor $\beta$ as: $\hat{F_i} = F_i + \beta\eta$, where $\eta \sim  \mathcal{N}(\eta \mid 0, \Sigma)$. Here, $\Sigma = \sigma^2 I$, and $\sigma$ is randomly sampled from a uniform distribution between 0 and 1 each time, making the covariance matrix $\Sigma$ dynamically change for every noise generation. Thus the final feature bank can be obtained by combining the original feature back and the augmented ones: $\mathcal{C}^{final} = \{(F_{i}^*, Y_{i}^*)\}_{i=1}^{m}$, where $F_i^* = \{F_i : i \in \{1, \dots, n\}\} \cup \{\hat{F}_{i-n} : i \in \{n+1, \dots, m\}\}, Y_i^* = \{Y_i : i \in \{1, \dots, n\}\} \cup \{Y_{i-n} : i \in \{n+1, \dots, m\}\}, m=2n$.

\subsection{Learning Objective}
One common challenge in deepfake detection is the imbalances in the training set. This is also evident in DFWild-Cup dataset, as shown in {Table~\ref{tab:dataset}.} To robustly handle imbalance issue and simultaneously improve model's performance on AUC metric, we 
leverage a distributionally robust optimization (DRO) technique called Conditional Value-at-Risk (CVaR) \cite{lin2024robust, lin2024robust2, lin2024preserving, ju2024improving, hu2024outlier, hu2023rank, hu2022distributionally, hu2021tkml, hu2022sum, hu2020learning,si2024meta,lin2024robust1,hu2024fairness,hu2022rank}, and integrate a binary vector-scaling (VS) loss~\cite{kini2021label} which designed for handling label imbalance. Additionally we add an AUC loss~\cite{yang2023minimax, lin2024robust3, wu2025preserving} to directly optimize the model's performance on the AUC metric. 

\textbf{Conditional Value-at-Risk (CVaR) Loss.} The CVaR loss is formulated to make the model focus on the hardest examples in the dataset, thereby improving robustness. It is defined as:

\begin{equation}
\mathcal{L}_{CVaR}(\theta) = \min_{\lambda \in \mathbb{R}} \left\{ \lambda + \frac{1}{\alpha m} \sum_{i=1}^{m} \left[ \ell(\theta; F_{i}^*, Y_{i}^*) - \lambda \right]_+ \right\},
\label{eq:cvar}
\end{equation}
where $\theta$ symbolizes the model parameters, and $(F_{i}^*,Y_{i}^*)$ are the feature-label pairs from the dataset. $[a]_+ = \max\{0, a\}$ is the hinge function. The parameter $\alpha \in (0,1]$ facilitates tuning the focus from the most challenging to more average examples. $\ell$ is the classification loss function (\eg, VS loss) to handle the imbalances, formulated as: $\ell = \omega_{Y_{i}^*} \cdot \log \left( 1 + e^{\Delta_{Y_{i}^*}} \cdot e^{-\zeta_{Y_{i}^*} Y_{i}^* h(F_{i}^*)} \right)$, where $\omega_{Y_{i}^*}$ is the weighting factor for class $Y_{i}^* \in \{0,1\}$. $h(F_{i}^*)$ is the predict logit. $\zeta_{Y_{i}^*}$ is the multiplicative logit scaling factor.  $\Delta_{Y_{i}^*}$ is the additive logit scaling factor. In practice, we set  $\zeta_{Y_{i}^*=0}=1.2$, $\zeta_{Y_{i}^*=1}=0.8$, $\Delta_{Y_{i}^*=0}=0.05$,  $\Delta_{Y_{i}^*=0}=-0.05$.

\textbf{AUC Loss}. To enhance the model's ability to distinguish between classes, especially under class imbalance, an area under the curve (AUC) loss is used. This loss directly optimizes the AUC metric by encouraging the model to rank positive examples higher than negative ones. It penalizes cases where positive samples are scored lower or equal to negative samples, helping the model better separate the two classes.  It is defined as:


\begin{equation}
\vspace{-1mm}
\mathcal{L}_{\text{AUC}} = \frac{1}{m^{+}m^{-}} \sum_{i=1}^{{m}^+} \sum_{j=1}^{{m}^-} \mathbb{I}_{[h(F_i^{*+}) \leq h(F_j^{*-})]},
\label{eq:auc_loss}
\end{equation}

where $F_i^{*+}$ means $F_i^{*}$ that has the positive label (\ie, $Y_i^{*}=1$) and $F_i^{*-}$ means $F_i^{*}$ that has the negative label (\ie, $Y_i^{*}=0$).  
In practice, the non-differentiable indicator function $\mathbb{I}$ can be replaced with a (sub)-differentiable and non-increasing surrogate loss function. For instance, in our experiments, we replace $\mathbb{I}_{[a\leq 0]}$ with the logistic loss $\log(1 + \exp(-a))$.

\textbf{Total Loss}. Therefore, the final learning objective becomes:

\begin{equation}
\vspace{-1mm}
\mathcal{L}(\theta) = L_{CVaR} + \gamma L_{AUC}, 
\label{eq:total_loss}
\end{equation}

where $\gamma$ is a trade-off hyperparameter that balances the contribution of the CVaR loss and the AUC loss. This learning objective is designed to address class imbalance robustly while simultaneously enhancing the model's performance in terms of the AUC metric.

\subsection{Optimization}
Last, to further improve the detector's generalization capability, we optimize the detection model by utilizing the sharpness-aware minimization (SAM) method~\cite{foret2020sharpness} to flatten the loss landscape. Note that this optimization module can be used in both supervised learning and semi-supervised learning. As shown in Fig.\ref{fig:SL}, by utilizing such a technique, the model yields a more flattened loss landscape indicating a stronger generalization capability~\cite{lin2024preserving, ren2024improving}. As a reminder, the model's parameters are denoted as $\theta$, flattening is attained by determining an optimal $\epsilon^*$ for perturbing $\theta$ to maximize the loss, formulated as:
\begin{equation}
    \begin{aligned}        
    \epsilon^*&=\arg\max_{\|\epsilon\|_2\leq \nu}{\mathcal{L}}\textbf{(}\theta+\epsilon \textbf{)}\\
    &\approx\arg\max_{\|\epsilon\|_2\leq \nu}\epsilon^\top\nabla_\theta \mathcal{L}=\nu\frac{\nabla_\theta \mathcal{L}}{\|\nabla_\theta \mathcal{L}\|_2},
    \end{aligned}
\label{eq:epsion_star}
\end{equation}
where $\nu$ is a hyperparameter that controls the perturbation magnitude. The approximation is obtained using first-order Taylor expansion with the assumption that $\epsilon$ is small. The final equation is obtained by solving a dual norm problem. As a result, the model parameters are updated by solving the following problem:
\begin{equation}
    \begin{aligned}
        \min_{\theta} \mathcal{L}\textbf{(}\theta+\epsilon^*\textbf{)}.
    \end{aligned}
\label{eq:sharpness}
\end{equation}
Perturbation along the gradient norm direction increases the loss value significantly and then makes the model more generalizable while detecting real and fake face images.

\subsection{Algorithm}

We first initialize the model parameters $\theta$ and then randomly select a mini-batch $\mathcal{C}_b$ from $\mathcal{C}^{final}$, performing the following steps for each iteration on $\mathcal{C}_b$ (see Algorithm~\ref{alg:Optimization}):

\begin{compactitem}
    \item Compute the VS loss $\ell(\theta_l;F_i^*,Y_i^*)$ for all instances in $\mathcal{C}_b$ and determine the optimal $\lambda$ via binary search to minimize $L_{\text{CVaR}}$ (Eq.~\ref{eq:cvar}).
    
    \item Evaluate $L_{\text{AUC}}$ (Eq.~\ref{eq:auc_loss}) and compute the total loss $\mathcal{L}(\theta)$ using Eq.~\ref{eq:total_loss}, then derive $\epsilon^*$ from Eq.~\ref{eq:epsion_star}.
    
    \item Update parameters via gradient descent at the perturbed point: $\theta\leftarrow\theta-\beta \nabla_\theta \mathcal{L}\big|_{\theta+\epsilon^*}$, where $\beta$ is the learning rate.
\end{compactitem}

The process repeats for \textit{max\_iterations} epochs to obtain the optimized parameters.

\begin{algorithm}[t!]
    \caption{End-to-end Training}\label{alg:Optimization}
    \begin{algorithmic}[1]
        \REQUIRE A training dataset $\mathcal{C}^{final}$ with size $m$, $\gamma$, $\alpha$, $\eta$, $\nu$, max\_iterations, num\_batch, learning rate $\beta$
        \ENSURE Optimized parameters $\theta$
        
        \STATE \textbf{Initialization:} $\theta_0$, $l=0$
        
        \FOR{$e=1$ to \textit{max\_iterations}}
            \FOR{$b=1$ to \textit{num\_batch}}
                \STATE Sample a mini-batch $\mathcal{C}_b$ from $\mathcal{C}^{final}$
                \STATE Compute $\ell(\theta_l;F_i^*,Y_i^*)$, $\forall (F_i^*,Y_i^*)\in \mathcal{C}_b$ using VS loss
                \STATE Use binary search to find $\lambda$ that minimizes $L_{CVaR}$ (Eq. \ref{eq:cvar}) on $\mathcal{C}_b$
                \STATE Compute $L_{AUC}$ (Eq. \ref{eq:auc_loss}) on $\mathcal{C}_b$
                \STATE Compute total loss $\mathcal{L}(\theta)$ based on  Eq.\ref{eq:total_loss}
                \STATE Compute $\epsilon^*$based on Eq. \ref{eq:epsion_star}
                \STATE Compute gradient approximation for the total loss
                \STATE Update $\theta$: $\theta_{l+1} \leftarrow \theta_l - \beta \nabla_\theta \mathcal{L} \big|_{\theta_l+\epsilon^*}$
                \STATE $l \leftarrow l+1$
            \ENDFOR
        \ENDFOR
        \STATE \textbf{return} $\theta_l$ \COMMENT{Return the optimized model parameters}
    \end{algorithmic}
\end{algorithm}

\section{Experiments}
\subsection{Experimental Settings}

\subsubsection{Datasets}
We use the training and test datasets~\cite{DeepfakeBench_YAN_NEURIPS2023} provided by the organizers for IEEE SPS Signal Processing Cup 2025 (DFWild-Cup) \cite{CUP}. These datasets comprise a compilation of publicly available resources designed for the DeepfakeBench evaluation, encompassing eight standard datasets: Celeb-DF-v1, Celeb-DF-v2, FaceForensics++, DeepfakeDetection, FaceShifter, UADFV, Deepfake Detection Challenge Preview, and Deepfake Detection Challenge. Detailed information of the dataset is shown in Table~\ref{tab:dataset}.
The data is utilized through a structured approach that involves loading, processing, and training models on both training and test datasets. 

\begin{table}[t]
\centering
\caption{\it \small Detailed information of the DFWild-Cup Dataset.}
\label{tab:dataset}
\scalebox{1.0}{
\begin{tabular}{c|c|c}
\hline
Dataset    & \# Real & \# Fake \\ \hline
Train      & 42,690  & 219,470 \\ \hline
Test       & 1,548   & 1,524   \\ \hline
\end{tabular}
}
\end{table}

\newcommand{\cmark}{\ding{51}} 
\begin{table*}[t!]
\centering
\caption{\it \small Test Set Performance Comparison}
\label{tab:validation_metrics}
\begin{tabular}{@{}c|c|c|c|c|c|c@{}}
\toprule
Method     & ACC      & AUC      & F1       & Precision & Recall   & EER      \\
\midrule
Baseline 1 \cite{cozzolino2024raising} &0.868815       &0.96418       &0.868530       &0.873008       &0.869232     & 0.108268      \\

Baseline 2 \cite{santosh2024robust} & 0.906901      & 0.964797      & 0.906868      & 0.907149       & 0.906809      & 0.096457      \\
\midrule
Ours       & \textbf{0.913737} & \textbf{0.969067} & \textbf{0.913693} & \textbf{0.914162}  & \textbf{0.913617} & \textbf{0.093176} \\
\bottomrule
\end{tabular}
\vspace{-2mm}
\end{table*}
\subsubsection{Evaluation Metrics}
This paper utilizes various evaluation metrics to measure the effectiveness of detection models using metrics like Area Under the Curve (AUC), accuracy, F1-score, precision, recall, and Equal Error Rate (EER). AUC evaluates the model’s capability to differentiate between classes across different thresholds, offering a holistic performance assessment. Accuracy denotes the ratio of correctly classified instances (both true positives and true negatives) to the total cases analyzed. The F1-score, calculated as the harmonic mean of precision and recall, serves as a balanced performance indicator, particularly beneficial for imbalanced datasets. Precision determines the proportion of correctly identified positive cases among all predicted positives, whereas recall reflects the proportion of actual positive instances that the model successfully detected. EER represents the point where false positive and false negative rates are equal, summarizing the trade-off between these two error types in a single metric\cite{rafique2023deep}.

\subsubsection{Baseline Methods}
We evaluate our methods against two baseline models that employ the CLIP architecture for feature extraction in detecting AI-generated images, focusing on robust deepfake detection with imbalanced data.

The first baseline method adopts the CLIP-based framework proposed by \cite{cozzolino2024raising}, which leverages multimodal (image-text) embeddings from a pretrained CLIP model \cite{radford2021learning} to distinguish AI-generated images. In line with their approach, we extract CLIP features and train a linear classifier on top of these embeddings, capitalizing on CLIP’s discriminative capabilities across diverse generative models.

The second baseline method follow the structure of the framework proposed by \cite{santosh2024robust}, preserving key components such as CLIP-based feature extraction, latent space augmentation, and sharpness-aware optimization. However, we replace the original cross-entropy loss with our VS loss for imbalance correction and substitute their AUC loss with our newly designed AUC loss to further enhance ranking performance. This adaptation isolates the impact of the improved loss formulations while maintaining the overall architecture.

\subsubsection{Implementation Details}
The AdamW optimizer is employed with a learning rate of 1e-5
and a weight decay of 6e-5 for effective regularization.
A Cosine Annealing Learning Rate Scheduler facilitates smooth convergence throughout the training process. The model trains for 100 epochs with a batch size of 256, using a seed of 8079 for reproducibility. The $\alpha$ and $\gamma$ are tuned on a grid search over the range \{0.1, 0.2, 0.3, 0.4, 0.5, 0.6, 0.7,0.8,0.9,1.0\} on the validation set. The optimal values are found to be $\alpha$ = 0.9 and $\gamma$ = 0.6. The $\nu$ in Eq.~\ref{eq:epsion_star} is set as 0.1. The best-performing model is selected based on the highest AUC score achieved during training.

\subsection{Results}
\begin{figure}[t]
  \centering
  \begin{subfigure}[t]{\linewidth}
    \centering
    \includegraphics[width=0.8\linewidth]{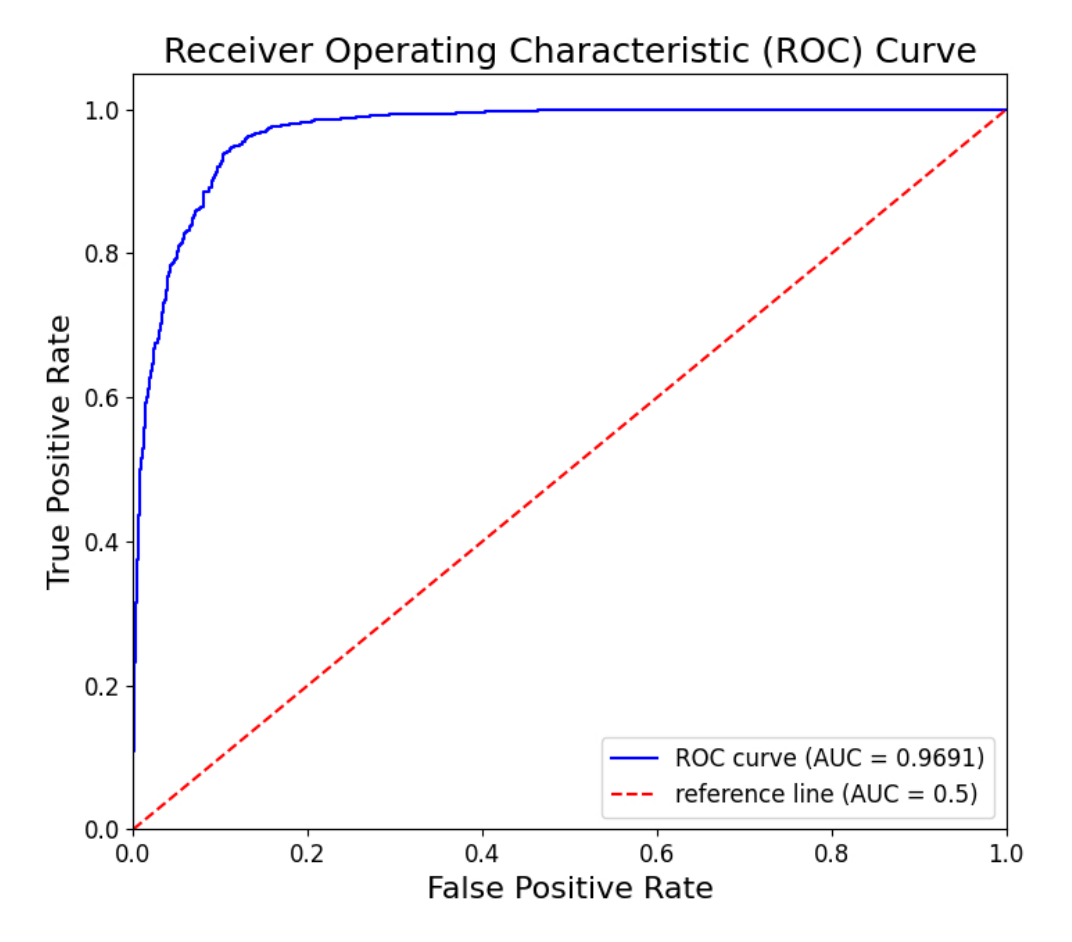}
    \vspace{-4mm}
    \caption{\it \small ROC curve of our proposed method.}
    \label{fig:roc}
  \end{subfigure}

  \vspace{4mm} 

  \begin{subfigure}[t]{\linewidth}
    \centering
    \includegraphics[width=1\linewidth]{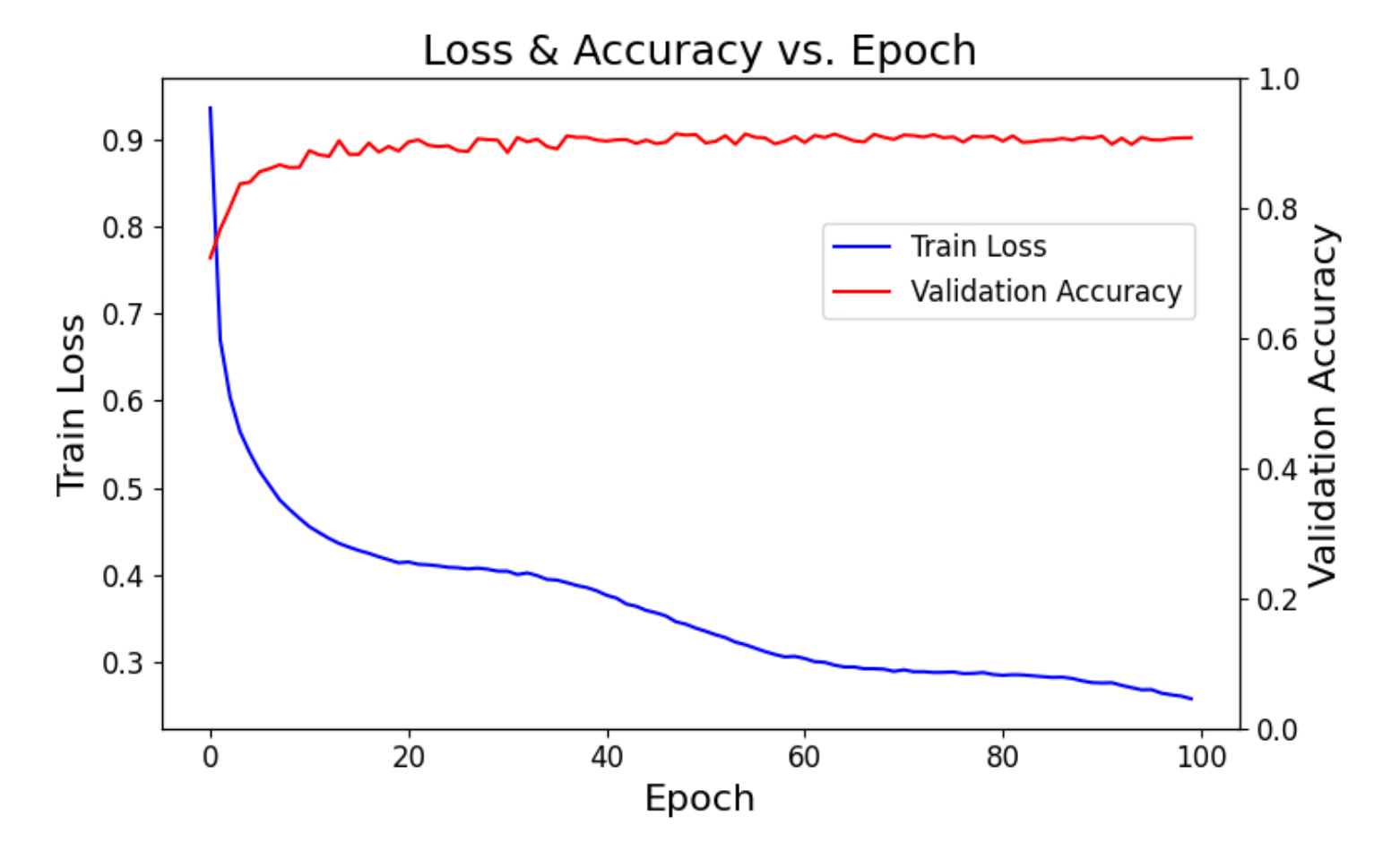}
    \vspace{-4mm}
    \caption{\it \small Training loss and validation accuracy over 100 epochs.}
    \label{fig:loss}
  \end{subfigure}

  \caption{\it \small Visualization of the ROC curve (a) and training dynamics (b) for our proposed method.}
  \label{fig:combined_fig}
\end{figure}

The performance of the proposed model on the test set demonstrates high performance across various metrics, as shown in Table~\ref{tab:validation_metrics}.
The model shows consistent improvements over the baselines across all evaluation metrics. Our method achieves the highest AUC of 0.969067, demonstrating superior discriminative capability compared to Baseline 1 and Baseline 2. Similarly, the F1 score of 0.913693 indicates strong balance between precision and recall, surpassing the baselines. Moreover, the model attains the highest accuracy (0.913737) and the lowest Equal Error Rate (EER) of 0.093176, further highlighting its robustness.  The validation performance is visually supported by the Receiver Operating Characteristic (ROC) curve in Fig.~\ref{fig:roc}, which demonstrates a near-perfect curve with an AUC close to 1, validating the model's discriminative capability. Additionally, Fig.~\ref{fig:loss} shows the progression of training loss and validation accuracy over 100 epochs, highlighting steady improvements and stable generalization throughout the training process.

\subsection{Ablation Study}

\textbf{Effect of Robust Loss.} We investigate the impact of CVaR loss by comparing the learning objective with combining Cross-Entropy (CE) loss and AUC loss using the same value of $\gamma$ for trading-off both terms. Table~\ref{tab:loss_ablation} shows that using $\gamma=0.9$ for the combined loss (CE + $\gamma$AUC) is worse than our proposed method (CVaR + $\gamma$AUC), which demonstrates the significance of CVaR loss for handling imbalanced data. 

\begin{table}[t!]
\centering
\caption{Ablation study of loss function composition}
\label{tab:loss_ablation}
\resizebox{\columnwidth}{!}{%
\begin{tabular}{l|c|c|c|c|c}
\toprule
Method & AUC & Accuracy & F1 & Precision & Recall \\
\midrule
CE + $\gamma$AUC ($\gamma=0.9$) & 0.964737 & 0.901367 & 0.901338 & 0.901549 & 0.901287 \\
\midrule
CVaR + $\gamma$AUC ($\gamma=0.9$) (Ours)       & \textbf{0.969067} & \textbf{0.913737} & \textbf{0.913693} & \textbf{0.914162}  & \textbf{0.913617} \\
\bottomrule
\end{tabular}
}
\vspace{-2mm}
\end{table}

\textbf{Impact of Model Architecture.} We examine how different MLP architectures affect performance when combined with CLIP features. Table~\ref{tab:arch_ablation} shows that increasing model depth beyond 3 layers leads to decreasing performance across all metrics. The 3-layer MLP achieves the best results (AUC=0.9530, Accuracy=0.8916), while deeper architectures suffer from significant performance degradation, with AUC dropping to 0.9446 for the 15-layer variant. This suggests that the CLIP features already provide strong representation power, and complex downstream models may lead to overfitting rather than improved performance.

\begin{table}[t!]
\centering
\caption{Ablation study of MLP architecture variants}
\label{tab:arch_ablation}
\resizebox{\columnwidth}{!}{%
\begin{tabular}{l|c|c|c|c|c}
\toprule
MLP Layers & AUC & Accuracy & F1 & Precision & Recall \\
\midrule
3 Layers & {0.9530610999} & \textbf{0.8916015625} & 
\textbf{0.8915974158} & \textbf{0.8915918935} & \textbf{0.8916076624} \\
6 Layers & \textbf{0.954188624} & 0.5960286458 & 0.5217020215 & 0.7708510168 & 0.599144947 \\
9 Layers & 0.952866538 & 0.4993489583 & 0.3387342288 & 0.7488569562 & 0.5032299742 \\
12 Layers & 0.949836424 & 0.500976563 & 0.3422753835 & 0.7492639843 & 0.5048449612 \\
15 Layers & 0.944614845 & 0.4996744792 & 0.3394440275 & 0.7489382555 & 0.5035529716 \\
\bottomrule
\end{tabular}
}
\vspace{-2mm}
\end{table}

\subsection{Sensitivity Analysis}
Table~\ref{tab:gamma_sensitivity} presents the performance metrics for different values of $\gamma$ in the CVaR(VS) + $\gamma$AUC objective. Our analysis demonstrates that $\gamma = 0.9$ achieves the most robust performance, attaining the highest Accuracy (0.908) and Precision (0.908) while maintaining competitive results across all other metrics: AUC (0.955), F1-score (0.907), and Recall (0.907). While $\gamma = 0.6$ yields a marginally higher AUC (0.962), this configuration shows reduced performance in classification metrics (Accuracy: 0.893, F1: 0.893). The model exhibits remarkable stability across the $\gamma$ spectrum, with F1 scores consistently ranging between 0.893--0.908. These results indicate that higher $\gamma$ values generally provide superior balance between ranking performance (AUC) and classification accuracy, with $\gamma = 0.9$ emerging as the optimal configuration that maintains predictive power while enhancing model robustness.

\begin{table}[t!]
\centering
\caption{Performance metrics across different $\gamma$ values}
\label{tab:gamma_sensitivity}
\resizebox{\columnwidth}{!}{%
\begin{tabular}{c|c|c|c|c|c}
\toprule
Metric &  $\gamma=0.5$ & $\gamma=0.6$ & $\gamma=0.7$ & $\gamma=0.8$ & $\gamma=0.9$ \\
\midrule
AUC  & 0.941155 & \textbf{0.962341} & 0.957956 & 0.941373 & {0.955130} \\
Accuracy  & 0.893555 & 0.892578 & 0.902344 & 0.897135 & \textbf{0.907552} \\
F1  & 0.893430 & 0.892574 & 0.902309 & 0.897061 & \textbf{0.907492} \\
Precision  & 0.894765 & 0.892804 & 0.902588 & 0.897784 & \textbf{0.908138} \\
Recall  & 0.893342 & 0.892678 & 0.902251 & 0.896982 & \textbf{0.907409} \\
\bottomrule
\end{tabular}
}
\vspace{-2mm}
\end{table}

\section{Conclusion}
The challenge of robust deepfake detection is compounded by severe class imbalance in real-world datasets and the increasing sophistication of generative models. To address these issues, we introduced a novel framework combining CLIP's powerful visual representations with a robust learning strategy featuring CVaR-based loss reweighting and AUC optimization. Our method dynamically prioritizes challenging samples while maintaining balanced learning across classes, further enhanced by sharpness-aware optimization for improved generalization. Extensive experiments on the DFWild-Cup benchmark demonstrate our approach's superiority, achieving state-of-the-art performance while significantly outperforming existing methods in imbalanced scenarios. The framework's effectiveness across diverse manipulation techniques and resilience to distribution shifts makes it particularly valuable for real-world deployment where data imbalances and evolving generation methods remain critical challenges.

\textbf{Limitation.} One limitation is that the empirical class distribution must be known in order to calculate the weighting factor in the VS loss. This requirement may be impractical when access to the full dataset is limited.

\textbf{Future Work.} We plan to evaluate our method to various datasets in the future and extend it to the detection for AI-generated voice \cite{ren2024improving}.

\smallskip
\smallskip
\noindent\textbf{Acknowledgments.} This work is supported by the U.S. National Science Foundation (NSF) under grant IIS-2434967 and the National Artificial Intelligence Research Resource (NAIRR) Pilot and TACC Lonestar6.
The views, opinions and/or findings expressed are those of the author and should not be interpreted as representing the official views or policies of NSF and NAIRR Pilot.

\bibliographystyle{ieeetr}
\bibliography{main}

\end{document}